# A Novel Apex-Time Network for Cross-Dataset Micro-Expression Recognition

Min Peng*
*Intelligent Security Center*
*Chongqing Institute of Green and Intelligent Technology*
Chongqing, China
pengmin@cigit.ac.cn

Chongyang Wang*
*UCL Interaction Centre*
*University College London*
London, United Kingdom
chongyang.wang.17@ucl.ac.uk

Tao Bi
*UCL Interaction Centre*
*University College London*
London, United Kingdom
t.bi@ucl.ac.uk

Yu Shi
*Intelligent Security Center*
*Chongqing Institute of Green and Intelligent Technology*
Chongqing, China
shiyu@cigit.ac.cn

Xiangdong Zhou
*Intelligent Security Center*
*Chongqing Institute of Green and Intelligent Technology*
Chongqing, China
zhouxiangdong@cigit.ac.cn

Tong Chen[†]
*College of Electronic and Information Engineering*
*Southwest University*
Chongqing, China
c_tong@swu.edu.cn

*Abstract*—The automatic recognition of micro-expression has been boosted ever since the successful introduction of deep learning approaches. As researchers working on such topics are moving to learn from the nature of micro-expression, the practice of using deep learning techniques has evolved from processing the entire video clip of micro-expression to the recognition on apex frame. Using the apex frame is able to get rid of redundant video frames, but the relevant temporal evidence of micro-expression would be thereby left out. This paper proposes a novel Apex-Time Network (ATNet) to recognize micro-expression based on spatial information from the apex frame as well as on temporal information from the respective-adjacent frames. Through extensive experiments on three benchmarks, we demonstrate the improvement achieved by learning such temporal information. Specially, the model with such temporal information is more robust in cross-dataset validations.

*Keywords—micro-expression, deep learning, neural network, feature fusion*

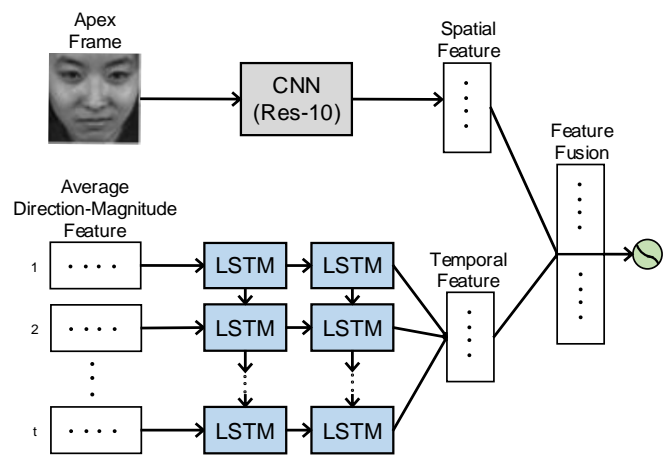

Fig. 1. An overview of ATNet.

## I. INTRODUCTION

Micro-expression exhibits when a person trying to suppress the underlying spontaneous emotion. It features low intensity and a brief period of time. Given the nature of micro-expression about revealing the hidden emotion, the recognition of it finds applications in many areas [1][2][3]. However, the automatic recognition of micro-expression is still difficult due to the characteristics of micro-expression, i.e. subtle facial movements and only lasting for a very short period of time.

Many efforts have been made to deal with automatic micro-expression recognition. First, researchers extracted sequence-based temporal features from an entire video clip of micro-expression for the recognition. The Local Binary Pattern histograms with Three Orthogonal Planes (LBP-TOP) was used in [4][5] to recognize the micro-expression. Then, another variant called LBP-SixIntersectionPoints (LBP-SIP) was used in [6]. Later, we also found the Riesz wavelet transform was employed for the representation extraction in [7], which is followed by fusion- and concatenation-based methods. More recently, [8] presented the Main Directional Mean Optical-flow (MDMO) method to acquire better a temporal representation. Another optical-flow based method called Facial Dynamics Map (FDM) was used in [9], which achieved a better recognition accuracy. Generally, the traditional methods for micro-expression recognition mainly focused on the temporal feature engineering originated from the video processing domain.

More related to our work are the studies based on deep learning. For many of the existing works [14][16], the common practice is to use neural networks like convolutional neural network (CNN) or Long-Short-Term-Memory network (LSTM) to extract spatial and temporal features in an end-to-end manner, and classification based on fully-connected layers would be performed in the end. As the micro-expression is only exhibited at transient moments during an experiment trail, applying neural networks on the entire video clip could be problematic. Inspired by Ekman's statement that 'snapshot taken at a point when the expression is at its apex can easily convey the emotion message' [10], some researchers [11][12][13][19] started to instead only use the apex frames for the recognition. The advantages of using apex frame is not only about getting rid of redundant information but also making it possible to transfer the knowledge learned from macro-expression recognition, where the available datasets mostly comprise images.

However, the trade-off between using apex frames and losing the temporal information is currently less mentioned.

In this paper, we propose a two-stream neural network, referred to as ATNet, with one stream of learning spatial feature from the apex frame and another stream of learning temporal feature from the respective-adjacent frames. The architecture of ATNet is shown in Fig. 1. Respectively, the spatial stream comprises Res-10 network [12] for spatial information extraction on the apex frame, while the temporal stream would extract temporal information with stacked vanilla LSTM network applied on the frames around the apex frame.

The contributions of this work can be summarized as:

- We propose a novel network architecture to combine the spatial feature learning on apex frame using CNN and the temporal feature learning on adjacent frames using LSTM network. To the best of our knowledge, this is the first work using spatio-temporal deep learning for cross-dataset micro-expression recognition.

- Extensive experiments are conducted on three benchmarks, namely CASME II [5], SAMM [24], SMIC [4]. In Composite Database Evaluation (CDE), where data from the three datasets are used together to do leave-one-subject-out validation (LOSO), the combination of spatial and temporal feature lead to even better accuracy. In Holdout-Database Evaluation (HDE), where the model is trained on two datasets and tested on the other one, our method is proved to be more robust than methods based on apex-frame. These imply the importance of including temporal information for cross-dataset micro-expression recognition.

## II. RELATED WORKS

### A. Deep Learning for Micro-Expression Recognition

Earlier studies on using deep learning techniques for the recognition of micro-expression mainly applied end-to-end neural networks like CNN and LSTM network on video clips containing micro-expressions. To name a few, first, Patel et al. [9] used CNN pre-trained on macro-expression databases to extract features from micro-expression clips. Then, Peng et al. [14] designed an end-to-end neural network with two streams in order to fit the two databases, namely CASME I [17] and CASME II [5], that originally have different framerates. Rrecently, Su-Jing et al. [16] further combined CNN with LSTM to do two-stages learning, where the spatial information of each frame within a clip is processed and passed to LSTM network for temporal information extraction. Indeed, the introduction of deep learning techniques so far has led to potent improvement in micro-expression recognition. However, due to the lack of samples and imbalance of categories [13], the practice of using entire video clip to learn the feature of micro-expression would easily lead to over-fitting. Therefore, approaches based on apex frame of micro-expression, where the facial-local movements reached its climax, are receiving more and more attention.

### B. Apex-Frame based Micro-expression Recognition

As micro-expression is transient, most frames of a micro-expression clip can be redundant for the recognition. To prove the efficiency of the recognition on the apex frame, Liong et al. [11] used Bi-WOOF method to extract optical-flow between onset and apex frame to recognize micro-expression and showed that this method outperformed many sequence-based approaches. Based on this finding learned on apex frame, several studies tried to adapt it for the development of neural networks. Peng et al. [12] used the single apex frame of each video clip to fine-tune a ResNet10 [12] that had been pre-trained with ImageNet and several macro-expression databases for the recognition of micro-expression. This method further won the first place in the Holdout-database Evaluation task at Micro-expression Grand Challenge 2018 [27]. In another study, Li et al. [13] used the magnified apex frames to fine-tune the VGG-FACE for the recognition and concluded that micro-expression recognition works well by only using apex frame information. More recently, another study done by Chongyang et al. [19] further adapted the attention mechanism [25] to the recognition of micro-expression on apex frames. The so-called micro-attention proposed in their work is able to learn to focus on the facial areas that contributed most to the recognition, while the results they achieved are also slightly better than [12].

However, it must be pointed out that the practices on apex frames would lead to the cost of losing dynamic temporal information. We believe the relevant temporal information can also contribute to the recognition of micro-expression. Furthermore, the generalization ability of a model trained on apex frames could be restricted to one dataset while using the temporal information is worth trying to improve this. In the next section, we will demonstrate how to combine the spatial information of apex frame with the temporal information surrounding it, through the engineering of deep learning architecture.

## III. METHODOLOGY

To take the advantage of the salient spatial information in apex frame while also to maintain the temporal information surrounding it, we propose a novel network named ATNet to implement such functions with an end-to-end manner. In this section, we first describe the data preparation where the apex frame along with the optical-flow based feature are generated. Then, we present the ATNet in detail.

### A. Average Direction-Magnitude feature

For CASME II [5] and SAMM [24], we directly use the apex frame indicated in the dataset. For SMIC [4] where the apex frame is not marked, we use the frame at the middle of each video clip as apex frame. Once the apex frame is located, we would extract the temporal information around it. To help the LSTM network learning informative temporal features from the micro-expression datasets, we compute a novel representation from the raw video sequences. The method proposed for the temporal representation extraction mainly adopted the optical-flow algorithm [18]. For the MDMO [8] and FDM [9] method, such algorithm has been largely modified according to the nature of micro-expression and further cooperated with complex alignment methods to achieve a better result. In [14], a vanilla optical-flow filed estimation method was adopted with a 3D-CNN. With the low-level feature and good spatial learning capability of CNN, this work achieved a state-of-the-art result back to then.

For our work, the temporal stream of ATNet is acted to provide auxiliary information for the decision making, which should better be computation-saving. Still, in order to provide enough information and to maintain a low complexity at the same time, a straightforward statistical optical-flow feature , referred to as average direction-magnitude feature, is designed. The pipeline of such temporal feature extraction process is shown in Fig. 2.

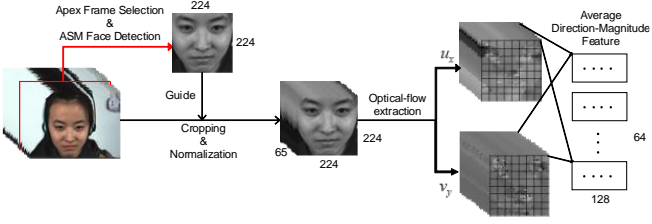

Fig. 2. The extraction of optical-flow feature from the frames adjacent to apex frame.

Given a video clip of micro-expression, we first locate the apex frame and then extract the facial area with ASM method [20]. For the clips where ASM fails, the extraction of face would be conducted manually. According to the facial area detected at apex frame, the facial area of frames within the same clip are extracted. Then, the size of all the processed frames within the clip are normalized to $224 \times 224$. To compute the optical-flow feature, 32 frames on the both side of the apex frame are used. Based on [14], given the selected 65 frames (including the apex frame) where each frame can be denoted as $I(x, y, t)$, the computation of optical-flow feature between the current frame and next frame after a lapse of time $\delta t$ can be written as:

$$I(x, y, t) = I(x + \delta x, y + \delta y, t + \delta t) \quad (1)$$

where $\delta x = u_x \delta t$, $\delta y = v_y \delta t$, with $u_x$ and $v_y$ to be the horizontal and vertical components that need to be estimated in the optical-flow field. According to [18], if the pixel value of frames within a clip is a continuous function of the spatial and temporal position, then we have:

$$I(x + \delta x, y + \delta y, t + \delta t) = I(x, y, t) + \delta x \frac{\partial I}{\partial x} + \delta y \frac{\partial I}{\partial y}$$
$$+ \delta t \frac{\partial I}{\partial t} + \varepsilon \quad (2)$$

where $\varepsilon$ is the two- or higher-order unbiased estimator of a lapse of time $\delta t$. When $\delta t$ approach the infinitesimal, the both sides of (2) can be divided by it and the computation of the optical-flow feature can be completed as:

$$\frac{\partial x}{\partial t}\frac{\partial I}{\partial x} + \frac{\partial y}{\partial t}\frac{\partial I}{\partial y} + \frac{\partial I}{\partial t} = 0 \quad (3)$$

which is

$$u\frac{\partial I}{\partial x} + v\frac{\partial I}{\partial y} + \frac{\partial I}{\partial t} = 0 \quad (4)$$

The matrix defined by $u_x$ and $v_y$, namely $[u_x^t, v_y^t]^T$, is the estimated optical-flow feature between two adjacent frames. For the 65 frames selected, 64 optical features can be respectively computed. For such optical-flow sequence, at spatial dimension, each frame is divided into $8 \times 8$ feature blocks, which we refer to as $B_t^i = [u_x^{t,i}, v_y^{t,i}]^T$, where $t =$ 1,2,...,64 is the index of feature maps and $i = 1,2,...,64$ is the index of blocks within a same feature map. Then, we transfer the optical-flow feature within each feature block $B_t^i$ into polar coordinates as $P_t^i(j) = [\rho_t^i(j), \theta_t^i(j)]^T$, where $j \in B_t^i$ is the spatial location at each feature block, $\rho$ and $\theta$ are the magnitude and direction attributes of current optical flow feature. Finally, we compute the average optical magnitude $\bar{\rho}_t^i$ and direction $\bar{\theta}_t^i$ for each feature block $B_t^i$, which can be written as:

$$\bar{\rho}_t^i = \frac{1}{J}\sum_{j=1}^{J}\rho_t^i(j), \quad j \in B_t^i \quad (5)$$

$$\bar{\theta}_t^i = \frac{1}{J}\sum_{j=1}^{J}\theta_t^i(j), \quad j \in B_t^i \quad (6)$$

Generally, for each of the 64 feature blocks at current time step t, the average optical magnitude $\bar{\rho}_t^i$ plus average direction $\bar{\theta}_t^i$ is computed. As a result, an average direction-magnitude feature vector with length of 128 is computed for each optical-flow frame. With the feature matrix of size $64 \times 128$ computed from all the frames around the apex frame in a video clip, the temporal stream of ATNet is able to better learn the dynamics of micro-expression and contribute to the final recognition.

*B. Apex-Time Network*

As shown in Fig. 1, the ATNet comprises two streams which are designed to combine the spatial learning on apex frame as well as the temporal learning on the adjacent frames. For the spatial stream, a ResNet-10 [12] network is used. Fig. 3 shows a classic residual architecture. To train the ResNet-10 for micro-expression recognition, we follow the idea raised in [12] [19], which means we firstly initialize it with ImageNet [26] then we do pre-training with four macro-expression benchmarks, namely Cohn-kanade dataset (CK+) [20], Oulu-CASIA NIR&VIS facial expression [21], Jaffe [22], and MUGFE [23]. Such pre-trained network is finally used as the spatial stream of ATNet.

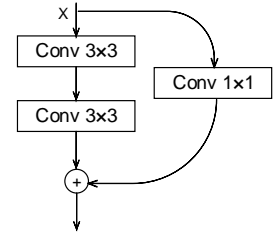

Fig. 3. Residual architecture.

For the temporal stream, a simple 2-layers vanilla LSTM network is used, which aims to reduce the overall complexity of ATNet. To cope with such simple network, the input has a slightly higher-level feature referred to as the average direction-magnitude feature of optical-flow field. Given a video clip, the input for the spatial stream of ATNet is the apex frame within the clip while 64 frames around the apex frame are used to compute the average direction-magnitude feature with size of $64 \times 128$. The dimension of the output vectors from both streams is 512. Before concatenating the two feature vectors, a L2 normalization is employed to the output of both streams, which can be written as:

$$C_i = \frac{O_i}{\sqrt{\sum_{d=1}^{D}O_i^d}} = \frac{O_i}{\sqrt{O_i^T O_i}} \quad (7)$$

where $O_i^d$ (i = 1,2; d = 1,2,...,512) is the output element from each stream, and $C_i$ is the normalized output. After such normalization, the feature vectors from both streams are concatenated to make the final vector with size of 1024 for classification.

## IV. EXPERIMENTS AND DISCUSSION

This section describes the dataset used and its pre-processing and the experimental results, being followed by an in-depth discussion about them.

### A. Data Preparation and Model Implementation

Three micro-expression benchmarks are used in our experiments, namely CASME II [5], SAMM [24] and SMIC [4]. Table 1 gives a summary of these three datasets. To employ these datasets for our experiment, we respectively merge the labels in CASME II and SAMM into the three categories, namely positive (Happiness), negative (Anger, Disgust, Sadness and Fear) and surprise according to SMIC. For the spatial-temporal normalization, apex-frame selection and temporal feature extraction, details can be found at Section 3. The training of both streams (spatial and temporal stream) of ATNet is conducted together. Here we need to mention that, to aid the spatial stream training where the number of apex frames is limited, a data augmentation strategy is used, namely rotation (with a maximum degree of 5) and pixel shift (with a maximum value of 10) under a selection probability of 0.5. Additionally, five macro-expression databases are used for pre-training to reduce the risk of over-fitting when dealing with the comparably smaller micro-expression datasets. We followed the procedure proposed in [12] [19] for the pre-processing of the five macro-expression datasets.

TABLE I. SUMMARY OF THE THREE MICRO-EXPRESSION BENCHMARKS

|  | Micro-Expression Datasets | | |
| --- | --- | --- | --- |
|  | CASME II [5] | SAMM [24] | SMIC [4] |
| Samples size | 255 | 159 | 164 |
| Participants Number | 35 | 32 | 16 |
| Resolution | 280 × 340 | 400 × 400 | 640 × 480 |
| Categories | Happiness, Surprise, Anger, Disgust, Sadness, Fear, others | Happiness, Surprise, Anger, Disgust, Sadness, Fear, others | Positive, Negative, Surprise |

The Caffe framework [27] is used in the implementation of ATNet. The number of hidden units used in each LSTM layer of the temporal stream of ATNet is 512. A drop-out layer with probability of 0.5 is used after the concatenation layer. The initial learning rate is 0.01 and it decreases 10 times smaller after every 10 epochs. The total epochs are 50. Momentum is set to 0.9 with weight-decay of 5e-6. For comparison, we also use the attention-based method [19] with the same input data of spatial stream of ATNet. The hyperparameter of all the neural network methods is tuned for best accuracy. In addition, two traditional methods, namely LBP-TOP and HOOF, using polynomial SVM (N=2) as classifier, are considered as the baselines. The input for these two traditional methods are the frames around the apex frame.

Two validation methods are used here, namely CDE and HDE. In CDE, data from the three datasets are used together to do LOSO. In HDE, a stricter cross-dataset validation is used, where two datasets act as training set and the rest one acts as testing set. For the recognition of the three micro-expression categories, we use Unweighted Average Recall (UAR) with Unweighted F1 Score (UF1) as the metric. The calculation of UF1 and UAR are written as:

$$\text{UF1} = \frac{1}{C}\sum_{c=1}^{C} \frac{2 \times TP_c}{2 \times TP_c + FP_c + FN_c} \quad (8)$$

$$\text{UAR} = \frac{1}{C}\sum_{c=1}^{C} ACC_c \quad (9)$$

$$ACC_c = \frac{TP_c}{N_c} \quad (10)$$

where $C$ is the number of classes, $TP_c, FP_c, FN_c$ are the number of the true positive, false positive and false negative respectively across all the folds of class $c$. $N_c$ is the number of samples under class $c$.

### B. Results and Discussion

The results achieved in CDE are summarized in Table 2. The CDE aims to test the generalization ability of a model across subjects in different datasets. For the results achieved by LBP-TOP, the best performance (Acc of 0.405, UF1 of 0.329 and UAR of 0.335) is seen under the smallest number of normalized frames (see results of TIM=10 VS. TIM=64). Under the same number of normalized frames, HOOF is able to produce better results (Acc of 0.475, UF1 of 0.350 and UAR of 0.350) than LBP-TOP (Acc of 0.378, UF1 of 0.293 and UAR of 0.294). Such results are owing to the fact that optical-flow feature is better at highlighting the subtle changes on the face than local statistical pattern. On the other hand, better results of HOOF are mostly achieved with larger number of normalized frames (TIM=64), which could be due to the dependency of the optical-flow feature on sequential frames. The Micro-attention method [19] that combined the apex frame with attention mechanism has generally better results comparing with the spatial stream of ATNet. Given the major difference of the two networks is the attention mechanism, the advantage of using which for the spatial pattern learning of micro-expression is proved. In the future, we may also implement the attention mechanism for the spatial stream of ATNet. Even better results are achieved by the temporal stream of ATNet, which implies the robustness of using the temporal feature for the recognition in such easier cross-dataset validation. The highest result among all the methods is achieved by ATNet (fusion) combining the apex and temporal information (the Acc, UF1 and UAR reach 0.693, 0.631 and 0.613 respectively), which are around 10% higher than using methods based on the apex frame. Once again, the importance of the temporal evidence of micro-expression is revealed.

The results achieved in HDE are summarized in Table 3. HDE is a stricter cross-dataset validation method than CDE, where an entire dataset is left out for testing. As is shown in the results, the apex-frame based methods (e.g., micro-attention and the spatial stream of ATNet) perform worse than the temporal stream of ATNet, which indicates that the model trained only with apex frames is not robust enough to go across different datasets. Although the proposed ATNet (fusion) has achieved better results than methods with the apex frame, the best results are acquired by the temporal

TABLE II. RESULTS FOR CDE VALIDATION

| Method (Bold: Best performance of each method at each column) | | Training and Testing Set | | | | | | | | | | |
|---|---|---|---|---|---|---|---|---|---|---|---|---|
| | | *Full (442)* | | | *CASME II (145)* | | | *SAMM (133)* | | | *SMIC (164)* | | |
| | | Acc | UF1 | UAR | Acc | UF1 | UAR | Acc | UF1 | UAR | Acc | UF1 | UAR |
| **LBP-TOP[a]** (on temporal frames) | {3,3,3,4} TIM=10 | **0.405** | **0.329** | **0.335** | 0.448 | 0.372 | 0.391 | 0.519 | 0.363 | 0.361 | 0.274 | 0.255 | 0.257 |
| | {3,3,3,8} TIM=10 | 0.396 | 0.293 | 0.294 | 0.441 | 0.285 | 0.288 | 0.511 | 0.301 | 0.302 | 0.262 | 0.246 | 0.246 |
| | {3,3,3,4} TIM=64 | 0.387 | 0.321 | 0.328 | 0.414 | 0.314 | 0.321 | 0.466 | 0.300 | 0.299 | **0.299** | **0.289** | **0.301** |
| | {3,3,3,8} TIM=64 | 0.378 | 0.305 | 0.305 | 0.421 | 0.302 | 0.302 | 0.444 | 0.289 | 0.288 | 0.287 | 0.281 | 0.279 |
| **HOOF[b]** (on temporal frames) | N=4, TIM=10 | 0.396 | 0.338 | 0.340 | 0.386 | 0.297 | 0.297 | 0.451 | 0.349 | 0.365 | 0.360 | 0.346 | 0.345 |
| | N=8, TIM=10 | 0.430 | **0.350** | **0.350** | 0.434 | 0.361 | 0.368 | 0.436 | 0.295 | 0.293 | **0.421** | **0.367** | **0.379** |
| | N=4, TIM=64 | **0.475** | 0.346 | **0.350** | 0.497 | 0.324 | 0.331 | 0.586 | **0.366** | **0.366** | 0.366 | 0.318 | 0.330 |
| | N=8, TIM=64 | **0.475** | 0.328 | 0.337 | **0.559** | **0.372** | **0.375** | **0.602** | 0.316 | 0.326 | 0.299 | 0.257 | 0.268 |
| **Micro-Attention [19]** (on apex frames) | | **0.613** | **0.508** | **0.493** | **0.703** | **0.539** | **0.517** | **0.677** | **0.403** | **0.340** | **0.482** | **0.473** | **0.466** |
| **ATNet** | Spatial Stream | 0.595 | 0.494 | 0.484 | 0.669 | 0.521 | 0.501 | 0.684 | 0.348 | 0.366 | 0.457 | 0.454 | 0.455 |
| | Temporal Stream | 0.624 | 0.581 | 0.581 | 0.724 | 0.691 | 0.679 | 0.632 | 0.487 | **0.490** | 0.530 | 0.527 | 0.530 |
| | Fusion | **0.693** | **0.631** | **0.613** | **0.834** | **0.798** | **0.775** | **0.701** | **0.496** | 0.482 | **0.561** | **0.553** | **0.543** |

a. {RXY,RXT,RYT,P} are the radius of the three planes of a video clip plus the number of neighboring points. TIM is the number of normalized frames.
b. N is the number of bins. TIM is the number of normalized frames.

TABLE III. RESULTS FOR HDE VALIDATION

| Method (Bold: Best performance of each method at each column) | | Testing Set (The other two is acted as training set) | | | | | |
|---|---|---|---|---|---|---|---|
| | | *CASME II (145)* | | *SAMM (133)* | | *SMIC (164)* | |
| | | UF1 | UAR | UF1 | UAR | UF1 | UAR |
| **LBP-TOP[c]** (on temporal frames) | {3,3,3,4} TIM=10 | 0.295 | 0.298 | 0.330 | **0.360** | 0.271 | **0.382** |
| | {3,3,3,8} TIM=10 | 0.240 | 0.249 | **0.335** | 0.347 | 0.235 | 0.315 |
| | {3,3,3,4} TIM=64 | **0.338** | **0.339** | 0.281 | 0.295 | 0.275 | 0.338 |
| | {3,3,3,8} TIM=64 | 0.313 | 0.317 | 0.248 | 0.263 | **0.289** | 0.334 |
| **HOOF[d]** (on temporal frames) | N=4, TIM=10 | **0.355** | **0.356** | 0.279 | 0.281 | 0.297 | 0.312 |
| | N=8, TIM=10 | 0.329 | 0.331 | 0.327 | **0.351** | **0.329** | **0.348** |
| | N=4, TIM=64 | 0.332 | 0.339 | **0.348** | **0.351** | 0.275 | 0.293 |
| | N=8, TIM=64 | 0.328 | 0.344 | 0.284 | 0.292 | 0.280 | 0.304 |
| **Micro-Attention [19]** (on apex frames) | | **0.338** | **0.369** | **0.383** | **0.380** | **0.354** | **0.372** |
| **ATNet** | Spatial Stream | 0.318 | 0.359 | 0.372 | 0.375 | 0.386 | 0.395 |
| | Temporal Stream | **0.631** | **0.643** | **0.450** | **0.458** | **0.503** | **0.524** |
| | Fusion | 0.523 | 0.501 | 0.429 | 0.427 | 0.497 | 0.489 |

c. {RXY,RXT,RYT,P} are the radius of the three planes of a video clip plus the number of neighboring points. TIM is the number of normalized frames.
d. N is the number of bins. TIM is the number of normalized frames.

stream of ATNet with an average 10% higher performance.

Therefore, based on the results of HDE that, we can conclude that, for the strict cross-dataset micro-expression recognition, using the apex frames for training has a high dependency on the original dataset and is not sufficient for the knowledge transferring from one dataset to another. Together with the results we have from CDE, we found that capturing the temporal evidence of micro-expression is able to make the modeling more robust across datasets.

Nevertheless, for CDE that can be deemed as an easier cross-dataset task, the apex frame is helpful (e.g., results achieved by the fusion version of ATNet is better than the temporal stream of it), especially if reducing the computation load during modelling is the priority to consider because modelling the apex frame does not need to consider the temporal dimension of the video data.

## V. CONCLUSION

For the cross-dataset micro-expression recognition, this work proposes a novel neural network architecture, named Apex-Time network (ATNet), that combines the spatial feature learned from apex frame with temporal feature learned from its adjacent frames. With two cross-dataset validation methods, namely composite database evaluation (CDE) and holdout-database evaluation (HDE), on the three micro-expression benchmarks (CASMEII, SAMM, SMIC), ATNet has shown better performance than other state-of-the-art methods. From the experiments we found that features learned from the apex frames are less transferable than the temporal feature learned from their adjacent frames. In HDE, the apex frame features learned from two datasets are even hindering the modelling performance on another dataset. Future work could focus on designing better temporal-dynamic feature extraction method for micro-expression.

ACKNOWLEDGMENT

This work is funded by the CAS Light of Western China Program, National Natural Science Foundation of China. (Grant No. 6180021609, Grant No. 6180070559 and Grant No. 61602433). This work is also funded by National key R&D program. (2018YFC0808303). Chongyang Wang is funded by UCL Overseas Research Scholarship and UCL Graduate Research Scholarship.

REFERENCES

[1] S. Porter, B. L. Ten, "Reading between the lies: identifying concealed and falsified emotions in universal facial expressions," Psychological Science, 2008, 19(5):508-514.

[2] T.A. Russell, E. Chu, M.L. Phillips, "A pilot study to investigate the effectiveness of emotion recognition remediation in schizophrenia using the micro-expression training tool," British Journal of Clinical Psychology, 2006, 45(4):579–583.

[3] S. Weinberger, "Airport security: Intent to deceive," Nature, 2010, 465(7297):412-5.

[4] X. Li, T. Pfister, X. Huang, G. Zhao, M. Pietikainen, A spontaneous micro-expression database: Inducement, collection and baseline, in: Automatic Face and Gesture Recognition, 2013, pp. 1–6.

[5] W.-J. Yan, S.-J. Wang, G. Zhao, X. Li, Y.-J. Liu, Y.-H. Chen, X. Fu, CASME II: An improved spontaneous micro-expression database and the baseline evaluation, PloS One 9 (2014) e86041.

[6] Y. Wang, J. See, C.W. Phan, et al, "LBP with Six Intersection Points: Reducing Redundant Information in LBP-TOP for Microexpression Recognition, "Computer Vision--Asian Conference on Computer Vision. Springer International Publishing, 2014:21–23.

[7] Y.H. Oh, A.C. Le Ngo, J. See, S.T. Liong, R.C.W. Phan, H.C. Ling, Monogenic Riesz wavelet representation for micro-expression recognition, in: Digital Signal Processing, IEEE, 2015, pp. 1237–1241.

[8] Y. J. Liu, J. K. Zhang, W. J. Yan, et al, "A Main Directional Mean Optical Flow Feature for Spontaneous Micro-Expression Recognition," IEEE Transactions on Affective Computing, 2016, 7(4):299-310.

[9] F. Xu, J. Zhang, J. Wang, "Micro-expression Identification and Categorization using a Facial Dynamics Map," IEEE Transactions on Affective Computing, 2017, 8(2): 254-267.

[10] P. Ekman, "Facial expression and emotion," American Psychologist, vol. 48, no. 4, pp. 384, 1993.

[11] S. T. Liong, J. See, K. S. Wong, and R. C. W. Phan, "Less is more: Micro-expression recognition from video using apex frame," Signal Processing: Image Communication, vol. 62, pp. 82–92, 2018.

[12] M. Peng, Z. Wu, Z. Zhang and T. Chen, "From Macro to Micro Expression Recognition: Deep Learning on Small Datasets Using Transfer Learning," 2018 13th IEEE International Conference on Automatic Face & Gesture Recognition (FG 2018), Xi'an, China, 2018, pp. 657-661.

[13] Y. Li, X. Huang, and G. Zhao, "Can Micro-Expression be Recognized Based on Single Apex Frame?", 2018 25th IEEE International Conference on Image Processing (ICIP), Athens, Greece, 2018, pp. 3094-3098.

[14] M. Peng, C. Wang, T. Chen, et al, "Dual Temporal Scale Convolutional Neural Network for Micro-Expression Recognition," Frontiers in Psychology, 2017, 8:1745.

[15] Gers, Felix A., Jürgen Schmidhuber, and Fred Cummins. "Learning to forget: Continual prediction with LSTM." (1999): 850-855.

[16] Wang, Su-Jing, Bing-Jun Li, Yong-Jin Liu, Wen-Jing Yan, Xinyu Ou, Xiaohua Huang, Feng Xu, and Xiaolan Fu. "Micro-expression recognition with small sample size by transferring long-term convolutional neural network." Neurocomputing 312 (2018): 251-262.

[17] Yan, W. J., Wu, Q., Liu, Y. J., Wang, S. J., and Fu, X. (2013). "CASME database: a dataset of spontaneous micro-expressions collected from neutralized faces," 2013 10th IEEE International Conference and Workshops on Automatic Face and Gesture Recognition (FG) (Shanghai).

[18] D. Sun, S. Roth, and M. J. Black, "A quantitative analysis of current practices in optical flow estimation and the principles behind them," International Journal of Computer Vision, vol. 106, no. 2, pp. 115–137, (2014).

[19] Wang, Chongyang, Min Peng, Tao Bi, and Tong Chen. "Micro-Attention for Micro-Expression recognition." arXiv preprint arXiv:1811.02360 (2018).

[20] P. Lucey, J. F. Cohn, T. Kanade, et al, "The Extended Cohn-Kanade Dataset (CK+): A complete dataset for action unit and emotion specified expression," Computer Vision and Pattern Recognition Workshops IEEE, 2010:94-101

[21] G. Zhao, X. Huang, M. Taini, et al, "Facial expression recognition from near-infrared videos," Image & Vision Computing, 2011, 29(9):607-619.

[22] M. Lyons, S. Akamatsu ,M. Kamachi, et al, "Coding Facial Expressions with Gabor Wavelets," Automatic Face and Gesture Recognition, 1998. Proceedings. Third IEEE International Conference on. IEEE, 1998: 200-205.

[23] N. Aifanti, C. Papachristou, A. Delopoulos, "The MUG Facial Expression Database," Proc 11th Int. Workshop on Image Analysis for Multimedia Interactive Services (WIAMIS), Desenzano, Italy, April 12-14, 2010.

[24] Davison, A. K., Lansley, C., Costen, N., Tan, K., & Yap, M. H., "SAMM: A spontaneous micro-facial movement dataset", IEEE Transactions on Affective Computing, 2016:9(1), 116-129.

[25] Itti, Laurent, and Christof Koch. "Computational modelling of visual attention." Nature reviews neuroscience. 2(3), 194 (2001).

[26] Deng, Jia, Wei Dong, Richard Socher, Li-Jia Li, Kai Li, and Li Fei-Fei. "Imagenet: A large-scale hierarchical image database." In Computer Vision and Pattern Recognition, (2009). CVPR 2009. IEEE Conference on, pp. 248-255. Ieee, 2009.

[27] Jia, Yangqing, Evan Shelhamer, Jeff Donahue, Sergey Karayev, Jonathan Long, Ross Girshick, Sergio Guadarrama, and Trevor Darrell. "Caffe: Convolutional architecture for fast feature embedding." In Proceedings of the 22nd ACM international conference on Multimedia, pp. 675-678. ACM, 2014.